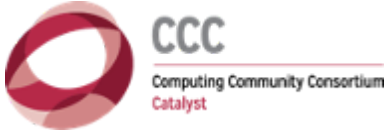

# Robotics Enabling the Workforce

*A Computing Community Consortium (CCC) Quadrennial Paper*

*Henrik Christensen (University of California San Diego), Maria Gini (University of Minnesota), Odest Chadwicke Jenkins (University of Michigan), and Holly Yanco (University of Massachusetts, Lowell)*

## Introduction

Robotics has the potential to magnify the skilled workforce of the nation by complementing our workforce with automation: teams of people and robots will be able to do more than either could alone. The economic engine of the U.S. runs on the productivity of our people. The rise of automation offers new opportunities to enhance the work of our citizens and drive the innovation and prosperity of our industries. Robotics research creates new options and possibilities for robotics technologies to be explored, evaluated, and transitioned. Most critically, we need research to understand how future robot technologies can best complement our workforce to get the best of both human and automated labor in a collaborative team. Investments made in robotics research and workforce development will lead to increased GDP, an increased export-import ratio, a growing middle class of skilled workers, and a U.S.-based supply chain that can withstand global pandemics and other disruptions. In order to make the United States a leader in robotics, we need to invest in basic research, technology development, K-16 education, and lifelong learning.

The use of robotics does not necessarily mean a reduction in jobs. An analysis of robot sales[1] and employment[2] over the last ten years shows a strong correlation between employment growth and robot sales. The economy in the US has grown significantly over the last decade; as an example, we have seen close to 1.1 million new jobs in manufacturing[2] and the growth in service sectors is even larger[3]. However, most of the growth has been in unskilled labor or labor with modest skills. Many jobs remain unfilled, even though they have higher salaries, because there is a lack of skilled labor available due to a lack of training for these jobs.

Robotics offers opportunities to solve real world problems while also creating new jobs. These new jobs include not only the use and maintenance of the robot systems, but their design and manufacture. It is critical to U.S. defense that the robot systems we depend upon be built in the U.S.; manufacturing these systems in the U.S. will also add jobs to our economy.

In addition to the desire to return manufacturing to the U.S., there are many additional benefits that robotics can provide to our society. For example, there is a move to provide new technologies in healthcare that improve quality of care without increased costs. As with manufacturing, the use of robotics in healthcare can increase worker productivity to keep costs low. Beyond healthcare, the age pyramid points to a need to assist people with daily tasks. There is an opportunity to leverage technology to deliver appropriate services without undue increases in cost and to increase the support for people who live alone or have no access to facilities for social interactions.

---

[1] https://ifr.org/downloads/press2018/WR%20Industrial%20Robots%202019_Chapter_1.pdf
[2] https://fred.stlouisfed.org/graph/?g=vcvN
[3] https://fred.stlouisfed.org/graph/?g=vcvX

Finally there is a tremendous need to upgrade the national infrastructure, from bridges and roads to dams and supply lines. In many cases safety and security are major challenges that can be addressed using robot technology, which also reduces the cost of delivering the services. There is no shortage of challenges in the U.S. where robotics and automation can assist to improve services, but only if paired with an educated set of operators.

**Opportunities for Robots to Impact Economic Growth**
The COVID-19 pandemic has revealed many needs that could be solved in part through the use of robotics, with sufficient investment in research, development, and job training. For example, in 2017 hospitality was a $1.6 trillion sector,[4] one that now has additional needs for disinfection, cleaning, maintenance, and socially distanced services. Similar needs are present in healthcare to provide safety to healthcare workers and patients alike. In agriculture, robotics could provide assistance where there is now a lack of seasonal workers to harvest crops. Robotics can address the supply chain issues that were revealed by COVID-19, including the need to manufacture critical PPE and other necessary items in the U.S. Autonomous trucks could increase the efficiency of our supply chain while also increasing road safety.

Robotics can assist with bringing manufacturing back to the U.S., as the addition of robots that work together with people increases worker productivity and thus offsets the higher labor costs in the U.S. In addition to shoring up the supply chain in the U.S., every new job in manufacturing generates 2.3 more jobs in the surrounding communities.[5]

**Catalyzing Societal Needs and Translation of Research into Economic Growth**
Many areas of our economy already rely on robots, but new opportunities and the improvement of existing systems can grow the U.S. economy. In addition to manufacturing, e-commerce relies extensively on robots to handle warehouses and distribution centers. Robots are also used to restock shelves in stores, keep track of inventory, and disinfect surfaces. Agriculture is relying on robots to increase productivity and reduce food cost. Improving the quality of life for the aging population is a major worldwide societal need, as well as supporting the education of children with developmental disabilities. Robots can enable people to do activities they could not do otherwise, and can provide social support in daily activities to people of all ages. To satisfy the societal needs mentioned above, there are multiple research challenges that need to be addressed at the research level and then translated into products that will produce economic growth.

---

[4] https://www.selectusa.gov/travel-tourism-and-hospitality-industry-united-states
[5] Andrew Liveris, Make it in America, Wiley & Sons, 2011

**Necessary Investments**

We need to invest in basic research, technology development, K-16 education, and lifelong learning to make the United States a leader in robotics and the developing areas described above. Investments in policy surrounding robotics is also necessary.

*Basic Research:* In order to make the fundamental discoveries that will allow robotics technology to make great leaps, we must invest in basic research across Federal agencies, including the Department of Agriculture, the Department of Defense, the Department of Energy, the Department of Transportation, the National Institutes of Health, the National Institutes for Standards and Technology, and the National Science Foundation. Not only does funding for fundamental research benefit robotics development, but it also trains the next generation of robotics researchers. We must also invest in interdisciplinary research in which robotics is a driving force to advance basic research in other fields.

*Technology Development:* To develop the economic benefits of a thriving robotics sector, investments are needed to move research findings from basic research into products. The expansion and replication of programs such as NSF's I-Corps will encourage recent graduates to form companies based upon the research they conducted as students. Through SBIR and STTR awards, we can grow the number of small companies working on robotics and create the foundations for the next round of large companies that will fuel job and economical growth.

*K-16 Education:* Key to developing a skilled workforce that will be able to design, develop, manufacture, operate, and maintain crucial robot systems is strengthening STEM education for all students from kindergarten through college. All too often, robotics programs are only offered in after school programs or in summer camps. We need to move from "opt in" programs that favor some children to programs that are offered to all students during the school day, so that we increase the number of people with strong STEM backgrounds and prepare a diverse workforce.

*Lifelong Learning:* We need to invest in lifelong learning for the U.S. population, by providing job training opportunities and opportunities to learn completely new skills. This will enable everyone to keep up with the exponential development of new technologies and to be qualified for the related job opportunities that will follow.

*Policy Development:* The U.S. needs to become a leader in the formulation and implementation of policies that put us at the forefront of robotics research and development. We have seen systems developed in the U.S. deployed first in other countries in the absence of a unified set of federal, state, and local policies that allow for robot systems (e.g., unmanned aerial systems, autonomous vehicles) to be deployed here.

*This white paper is part of a series of papers compiled every four years by the CCC Council and members of the computing research community to inform policymakers, community members and the public on important research opportunities in areas of national priority. The topics chosen represent areas of pressing national need spanning various subdisciplines of the computing research field. The white papers attempt to portray a comprehensive picture of the computing research field detailing potential research directions, challenges and recommendations.*

*This material is based upon work supported by the National Science Foundation under Grant No. 1734706. Any opinions, findings, and conclusions or recommendations expressed in this material are those of the authors and do not necessarily reflect the views of the National Science Foundation.*

*For citation use: Christensen H., Gini M., Jenkins C., & Yanco H. (2020) Robotics Enabling the Workforce. https://cra.org/ccc/resources/ccc-led-whitepapers/#2020-quadrennial-papers*